\documentclass{article}
\usepackage[utf8]{inputenc}
\usepackage{authblk}
\usepackage{setspace}
\usepackage[margin=1.25in]{geometry}
\usepackage{graphicx}
\graphicspath{ {./figures/} }
\usepackage{subcaption}
\usepackage{amsmath}
\usepackage[ruled,vlined]{algorithm2e}
\usepackage{soul}

\usepackage[style=nejm, 
citestyle=numeric-comp,
sorting=none]{biblatex}
\date{}

\begin{document}
\title{Active learning for efficient annotation in precision agriculture: a use-case on crop-weed semantic segmentation}

\author[1,*]{\small Bart M. van Marrewijk}
\author[2]{Charbel Dandjinou}
\author[1]{Dan Jeric Arcega Rustia}
\author[2]{Nicolas Franco Gonzalez}
\author[2]{Boubacar Diallo}
\author[2]{Jérôme Dias}
\author[2,3]{Paul Melki}
\author[1]{Pieter M. Blok}

\affil[1]{Wageningen University \& Research, Wageningen, The Netherlands}
\affil[2]{Exxact Robotics, Épernay, France}
\affil[3]{IMS, CNRS, University of Bordeaux, France}
\affil[*]{Corresponding author. Email: bart.vanmarrewijk@wur.nl}

\maketitle

\begin{abstract}
\noindent Optimizing deep learning models requires large amounts of annotated images, a process that is both time-intensive and costly. Especially for semantic segmentation models in which every pixel must be annotated. A potential strategy to mitigate annotation effort is active learning. Active learning facilitates the identification and selection of the most informative images from a large unlabelled pool. The underlying premise is that these selected images can improve the model's performance faster than random selection to reduce annotation effort. While active learning has demonstrated promising results on benchmark datasets like Cityscapes, its performance in the agricultural domain remains largely unexplored. This study addresses this research gap by conducting a comparative study of three active learning-based acquisition functions: Bayesian Active Learning by Disagreement (BALD), stochastic-based BALD (PowerBALD), and Random. The acquisition functions were tested on two agricultural datasets: Sugarbeet and Corn-Weed, both containing three semantic classes: background, crop and weed. Our results indicated that active learning, especially PowerBALD, yields a higher performance than Random sampling on both datasets. But due to the relatively large standard deviations, the differences observed were minimal; this was partly caused by high image redundancy and imbalanced classes. Specifically, more than 89\% of the pixels belonged to the background class on both datasets. The absence of significant results on both datasets indicates that further research is required for applying active learning on agricultural datasets, especially if they contain a high-class imbalance and redundant images. Recommendations and insights are provided in this paper to potentially resolve such issues.

\end{abstract}

\section{Introduction}

Deep learning has received much attention in the agricultural domain due to its potential for resolving complex computing tasks \cite{kamilaris2018}. Within the deep learning subdomain, semantic segmentation stands out for its pivotal role in precisely distinguishing class-specific pixels within agricultural imagery \cite{luo2023}. Semantic segmentation can be used for all kinds of precision-related crop monitoring tasks and targeted crop interventions. However, to achieve accurate performance, the segmentation model must be trained on a significant number of annotated images, a process that is time-consuming and expensive \cite{li2023labelefficient}. To address this challenge, active learning has emerged as a promising approach to optimize the annotation process by actively selecting the most informative images for annotation \cite{gal2017}. Selection of those informative images is typically done by uncertainty-based acquisition functions, which acquire images about which the segmentation model is most uncertain. By selecting these uncertain images and using them to retrain the segmentation model, performance can be significantly improved with minimal annotation effort. Thus, active learning is a promising solution to reduce annotation effort and improve efficiency in precision farming applications \cite{zahidi2021, blok2022active}.

One of the remaining challenges in active learning is the design and use of uncertainty-based acquisition functions. A crucial part in designing acquisition functions is the concept of aleatoric and epistemic uncertainty. Aleatoric uncertainty refers to the inherent randomness or variability in the data itself \cite{jospin2022}. It includes the uncertainty that cannot be reduced even if we had access to more annotated images or improved models. Epistemic uncertainty, on the other hand, refers to the uncertainty of the model arising from a lack of knowledge or limited data \cite{jospin2022}. Epistemic uncertainty can be reduced with more annotated images and it is therefore crucial in active learning because it provides essential information to select the most informative images. Acquisition functions that incorporate epistemic uncertainty are therefore preferred over functions that incorporate aleatoric uncertainty. One of the acquisition functions that only work with epistemic uncertainty is Bayesian Active Learning by Disagreement (BALD) \cite{houlsby2011bayesian}. Other hybrid functions of BALD, such as PowerBALD, combines the concept of epistemic uncertainty and stochastic sampling \cite{kirsch2022stochastic}. BALD, and to a lesser extent PowerBALD, have been applied to well-known public datasets such as Cityscapes, but to the best of our knowledge, these methods have not been tested for semantic segmentation on agricultural datasets. As these datasets are often not as diverse and balanced as other public datasets, the question is whether uncertainty-based active learning is still able to improve annotation efficiency on agricultural datasets. 

The aim of this paper is to bridge this research gap through a comparative study between BALD, PowerBALD and Random acquisition functions on agricultural image datasets. Our goal is to provide insights into the challenges and opportunities of active learning by testing different acquisition functions on these datasets. We have conducted such experiments on two crop-weed datasets, one of which is publicly available \cite{chebrolu2017ijrr} and the other was used in a practical production setup. The contributions of our work include:  1.) Detailed analysis of how active learning, especially uncertainty sampling, works on agricultural datasets; 2.) Identification of issues arising from applying active learning; and 3.) Potential solutions to resolve such issues.

\vspace{-2mm}

\section{Active learning}

    

%

The proposed active learning framework is based on uncertainty sampling (BALD) and stochastic uncertainty sampling (PowerBALD). Both methods select images the semantic segmentation model is most uncertain about, with the assumption that these images will contribute most to the performance when retraining the segmentation model. In this work, the level of uncertainty on each unlabeled image was determined with Monte-Carlo dropout, a frequently used sampling technique in active learning \cite{gal2016}. With Monte-Carlo dropout, unlabeled images were repeatedly passed through the model in inference mode with dropout activated, thus producing multiple segmentations for the same image. If the outputs for the same image show large variation, this may indicate that the model is uncertain about this image, making it a candidate for selection and annotation. This approach is presented in more detail in the sections below. First, an introduction to the active learning framework is presented in Section~\ref{sec:approach}, then the semantic segmentation model used in this work is explained Section~\ref{sec:segment_model}. In Section~\ref{sec:uncertainty} the different uncertainty acquisition functions are presented.

\vspace{-2mm}

\subsection{Active learning framework}\label{sec:approach}
Our active learning framework was developed using the Bayesian Active Learning (Baal) software, which has been specially developed for active learning \cite{atighehchian2019baal}. The pseudo-code is given in Algorithm \ref{algo1}. In the text below, the main components of our active learning framework are described in \textit{italics}.

Our active learning process starts with an \textit{initial dataset} randomly sampled from the \textit{training pool}. The training pool is the dataset with unlabeled images from which the active learning framework selects the most uncertain images. A semantic segmentation model is trained on this initially sampled dataset. After training, the model is used to calculate the uncertainty of each image in the training pool, using one of the acquisition functions, which are described in Section~\ref{sec:uncertainty_sampling}. Depending on the \textit{number of Monte-Carlo iterations}, each image is \textit{x} times inferred through the model with dropout enabled. After each iteration, the predicted model output for every pixel may be different, since the model is forced to make its decision with a potentially different set of model weights due to the dropout. The dropout intensity was made adjustable through the \textit{dropout probability}, whose value could be set between 0 (no dropout) and 1 (complete dropout). From the multiple model predictions, the uncertainty is calculated based on the \textit{acquisition function}. This outputs a list of the most informative images ready to be annotated. After annotating these images the segmentation model was trained again on the combined set of initial images and most informative images. This loop was repeated depending on the \textit{number of iterations}. This process iteratively increases the number of annotated images and thereby the performance of the segmentation model.



\SetAlFnt{\small}
\SetAlCapFnt{\small}
\SetAlCapNameFnt{\small}
\begin{algorithm}[h]
\caption{Active learning}
\label{algo1}

\SetKwData{SampIter}{numberOfIterations}
\SetKwData{InitDSSize}{initialDatasetSize}
\SetKwData{Poolsize}{sampleSize}
\SetKwData{Trainpool}{trainingPool}
\SetKwData{Valset}{valSet}
\SetKwData{InitDS}{initialDataset}
\SetKwData{TrainModel}{model}
\SetKwData{mIoU}{mIoU}
\SetKwData{AvImg}{availableImages}
\SetKwData{Pool}{pool}
\SetKwData{FP}{numberOfMonteCarloIterations}
\SetKwData{Method}{acquisitionFunction}
\SetKwData{AnnotTrain}{annotatedTrainSet}
\SetKwData{AnnotPool}{annotatedPool}
\SetKwData{DPROB}{dropoutProbability}
\SetKwData{EmptyArray}{[ ]}
\SetKwData{bracketopen}{(}
\SetKwData{bracketclose}{)}

\SetKwFunction{InitDSFunc}{Random}
\SetKwFunction{TrainModelFunc}{\hl{TrainModel}}
\SetKwFunction{EvalFunc}{\hl{EvaluateModel}}
\SetKwFunction{Sampling}{\hl{Sampling}}
\SetKwFunction{Annotate}{\hl{Annotate}}

\SetKwInOut{Input}{Inputs}
\SetKwInOut{Output}{Outputs}
\Input{acquisitionFunction : BALD, PowerBALD, Random \newline numberOfIterations : integer \newline sampleSize : integer \newline initialDatasetSize : integer \newline dropoutProbability : float \newline numberOfMonteCarloIterations : integer \newline trainingPool : dataset with all available images for training and sampling \newline valSet : dataset with images for validation}
\Output{mIoU : semantic segmentation performance \newline}

\SetKwFunction{FMain}{ActiveLearning}
\SetKwProg{Fn}{Function}{:}{}
\Fn{\FMain{\Method, \SampIter, \Poolsize, \InitDSSize, \DPROB, \FP, \Trainpool, \Valset}}
    {
        \mIoU$\gets$ \EmptyArray\;
        \InitDS$\gets$ \InitDSFunc{\Trainpool, \InitDSSize}\;
        \AnnotTrain $\gets$ \Annotate{\InitDS}\;
        \AvImg$\gets$ \textit{\Trainpool} $-$ \AnnotTrain\;
        \TrainModel$\gets$ \TrainModelFunc{\AnnotTrain}\;
        \mIoU.insert\bracketopen\EvalFunc{\Valset}\bracketclose\;
        
        \For{$i \gets 1$ \KwTo \SampIter}{
            {\Pool$\gets$ \Sampling{\TrainModel, \AvImg, \Method, \FP, \Poolsize, \DPROB}\;}
            \AnnotPool$\gets$ \Annotate{\Pool}\;
            \AnnotTrain$\gets$ \AnnotTrain $+$ \AnnotPool\;
            \TrainModel$\gets$ \TrainModelFunc{\AnnotTrain}\;
            \mIoU.insert\bracketopen\EvalFunc{\Valset}\bracketclose\;
            \AvImg$\gets$ \AvImg $-$ \AnnotPool\;
        }
        \KwRet \mIoU   
    }
\end{algorithm}

\vspace{-2mm}

\subsection{Segmentation model}\label{sec:segment_model}
\begin{figure}[h]
\centering
\includegraphics[width=.55\linewidth]{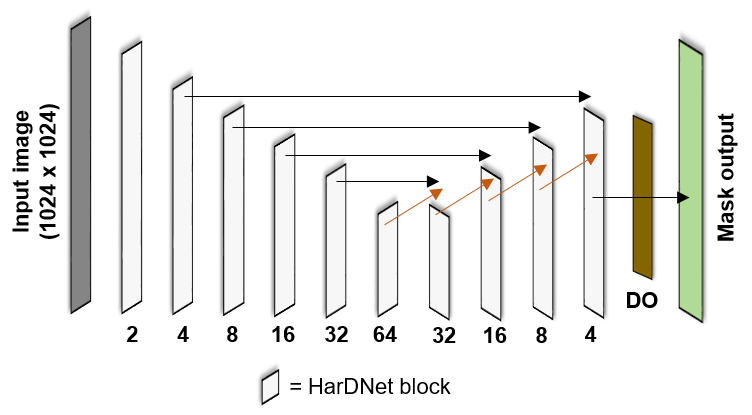}
\caption{FCHarDNet network architecture \cite{chao2019fchardnet} with dropout (DO). The numbers in the figure indicate the number of strides for a given block.}
\label{fig:fchardnet_architecture}
\end{figure}

Our active learning framework could potentially work with any segmentation model as long as there is a dropout layer added to that model. In our experiments, the Fully Convolutional HarDNet (FCHarDNet) was used as semantic segmentation model. FCHarDNet is a semantic segmentation model that employs a harmonic densely connected network \cite{chao2019fchardnet}, of which its architecture is shown in Figure \ref{fig:fchardnet_architecture}. Compared to other segmentation models, FCHardNet has a very low memory traffic, allowing it to have a very low inference time, which is relevant for potential real-time segmentation of agricultural images containing crops and weeds. To perform active learning, a dropout (DO) layer was added before the final convolutional layer of the FCHarDNet architecture (Figure \ref{fig:fchardnet_architecture}). By adding this dropout layer, Monte-Carlo iterations could be performed. If an image is relatively new to the currently trained model, we expect the pixel uncertainty to be relatively high when applying multiple Monte-Carlo iterations. We hypothesized that images with high model uncertainty contributed the most to the segmentation performance when retraining FCHarDNet.

\vspace{-2mm}

\subsection{Acquisition functions}\label{sec:uncertainty}
\label{sec:uncertainty_sampling}
After obtaining multiple model predictions with Monte-Carlo dropout, the predictions were summarized by a single value that represents the model uncertainty on that image. In our research, three different acquisition functions were tested: BALD\cite{houlsby2011bayesian}, PowerBALD \cite{kirsch2022stochastic}, and Random acquisition.
\begin{itemize}
\item BALD: BALD was calculated using Equation~\ref{eq:BALD}, which was obtained from \cite{gal2017,houlsby2011bayesian}.  The first term of BALD, the marginal entropy $H(y | x, D)$, triggers on images with high entropy. These are images that contain large fractions of pixels with distributed prediction probabilities across all classes (indicating that the model is not able to make a confident prediction on a single class per pixel). The second term of BALD, the conditional uncertainty $E_{p(\omega|D)} [H(y|x, \omega, D)]$, penalizes images with aleatoric uncertainty. Because of BALD's two terms, we were able to get better estimates of epistemic uncertainty rather than aleatoric uncertainty. 



\begin{equation}
{
\label{eq:BALD}
  BALD(y; \omega | x, D) = H(y | x, D)
  - E_{p(\omega|D)} [H(y|x, \omega, D)]
}
\end{equation}

  
\item PowerBALD: A limitation of using BALD is that it can sample a set of uncertain images with large visual similarities between them. This can especially occur with agricultural datasets that are obtained with slow-moving platforms, because then there can be a high visual similarity between the image at time \textit{t} and the image at time \textit{t}+1. Instead of relying on the single-most uncertain images from the training pool when using BALD, we also used PowerBALD \cite{kirsch2022stochastic}. PowerBALD uses stochastic sampling to sample a more diverse set of images from the uncertainty distribution of BALD. PowerBALD uses a temperature factor \textit{T} to conduct this sampling (Equation \ref{eq:PowerBALD}). In our experiments, \textit{T} was set to 1.0.

\begin{equation}
{
\label{eq:PowerBALD}
  PowerBALD(y; \omega | x, D) = \frac{BALD(y; \omega | x, D) ^ \frac{1}{T}}{\sum BALD(y; \omega | x, D)}
}
\end{equation}



\item Random acquisition: This approach selects images randomly from the training pool. It is used as the default method for comparison with BALD and PowerBALD.
\end{itemize}

\vspace{-6mm}
\section{Datasets}
\label{sec:dataset}
\begin{figure}[h]
\centering
    \begin{subfigure}[b]{0.35\linewidth}
        \centering
        \label{fig:sugarbeet_example}
        \includegraphics[width=\textwidth]{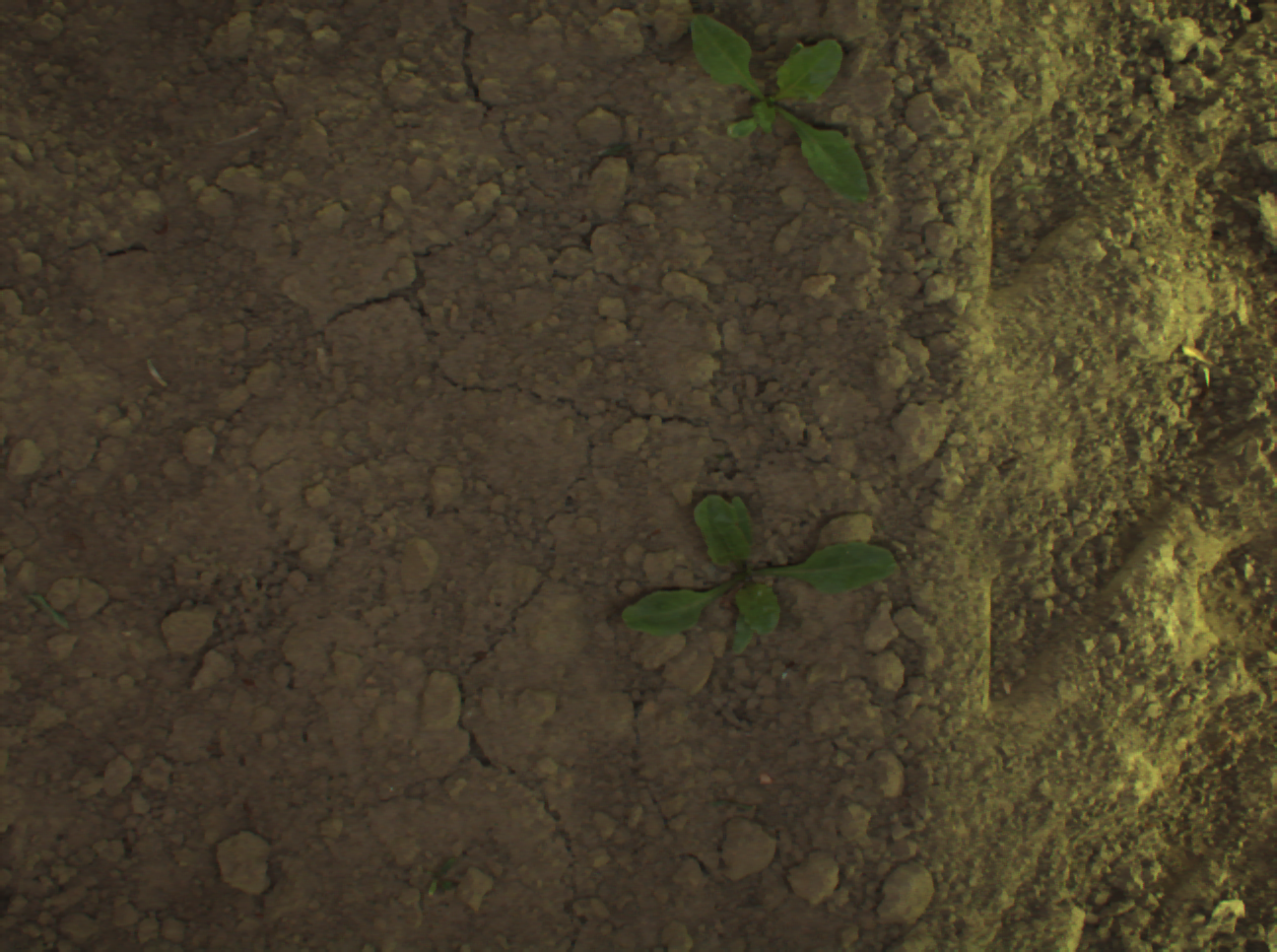}
        \caption{}
    \end{subfigure}
    \begin{subfigure}[b]{0.35\linewidth}
        \centering
        \includegraphics[width=\textwidth]{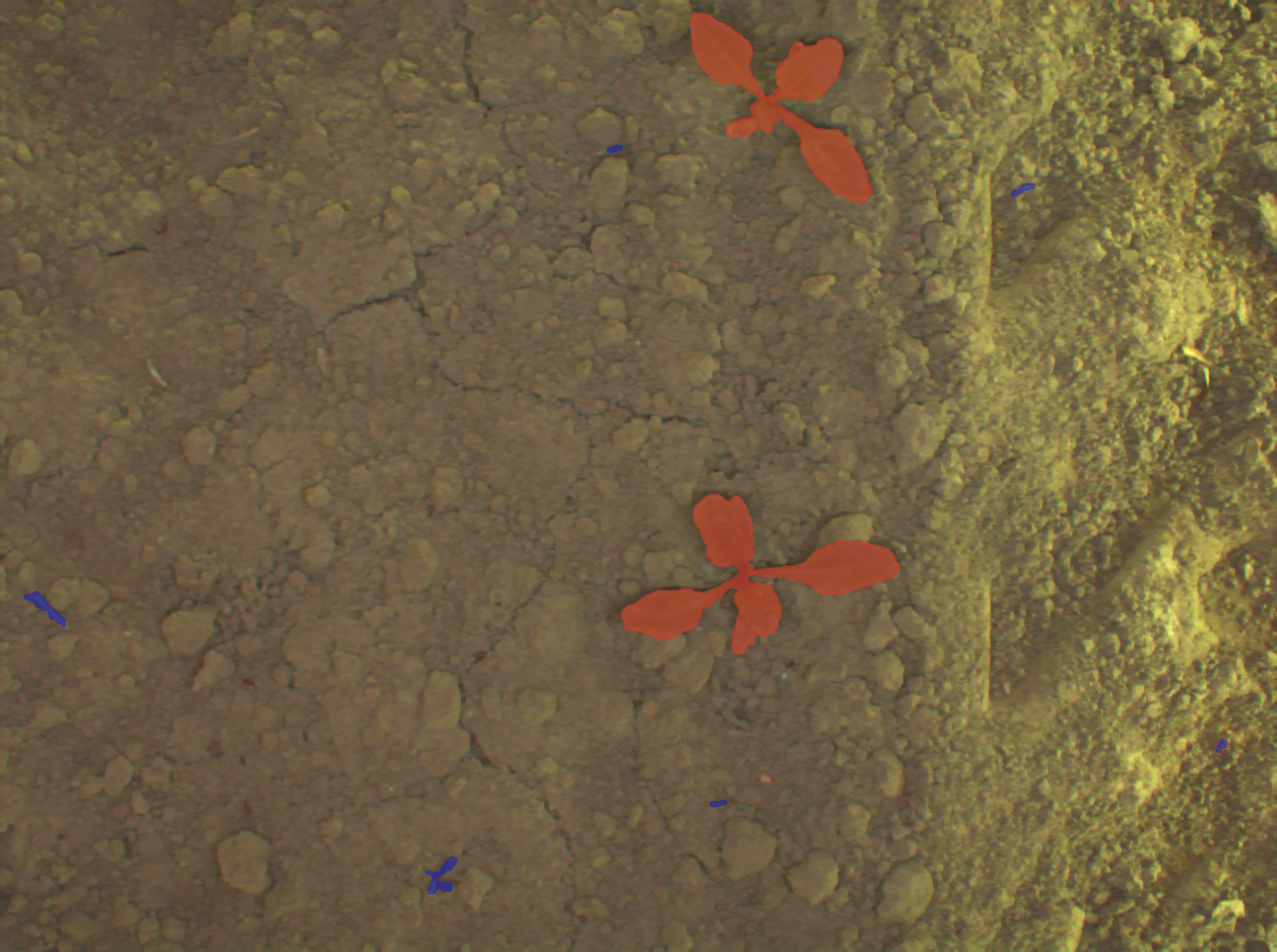}
        \caption{}
    \end{subfigure}
    \begin{subfigure}[b]{0.35\linewidth}
        \centering
        \label{fig:sugarbeet_example}
        \includegraphics[width=\textwidth]{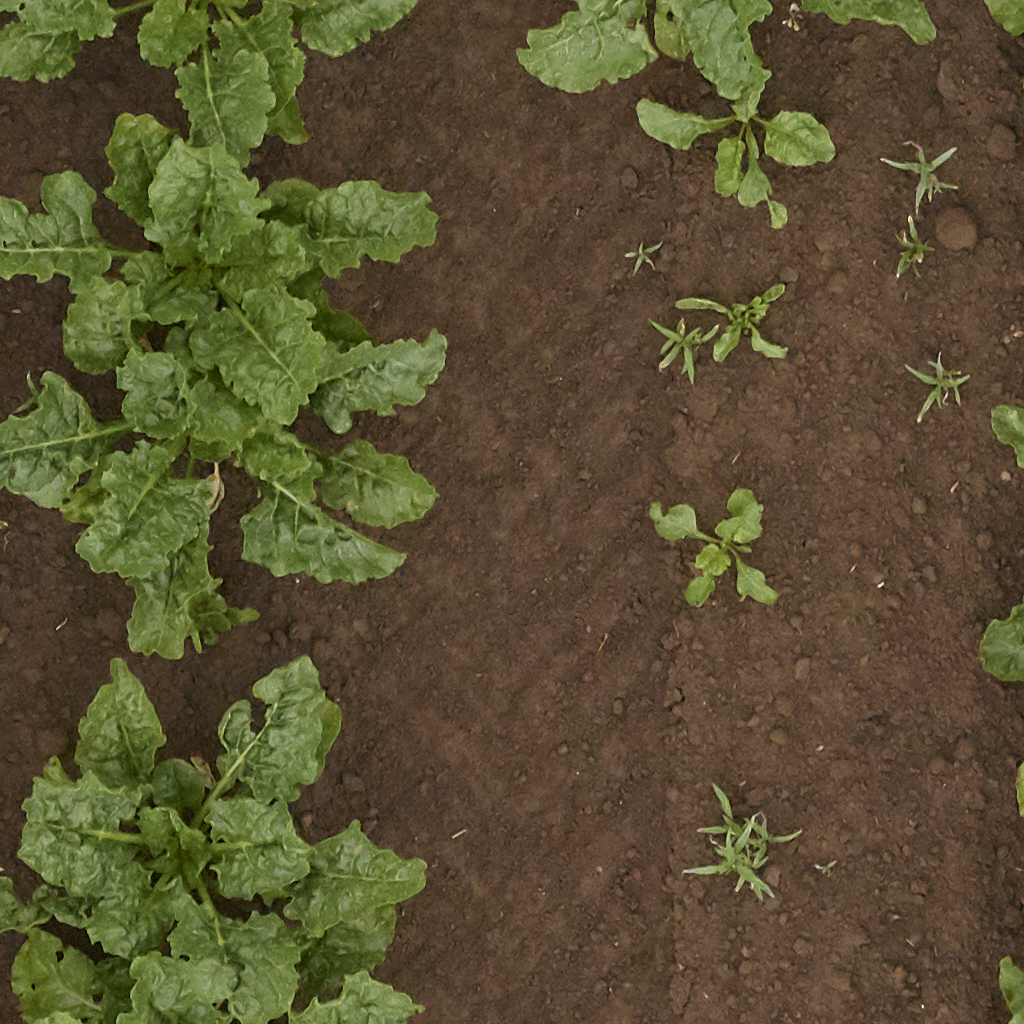}
        \caption{}
    \end{subfigure}
    \begin{subfigure}[b]{0.35\linewidth}
        \centering
        \includegraphics[width=\textwidth]{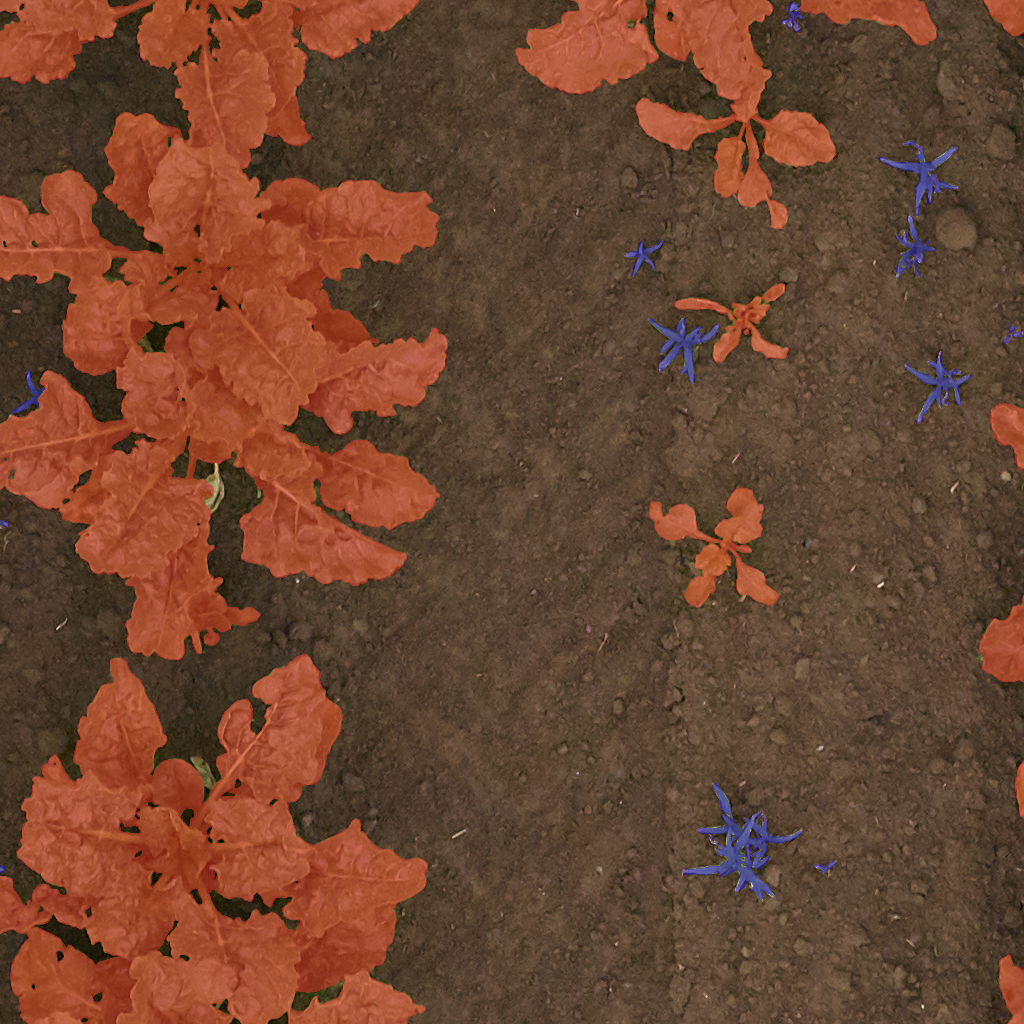}
        \caption{}
    \end{subfigure}
\caption{Example image from the Sugarbeet dataset with (a) input image and (b) corresponding annotation, where red=sugarbeet, blue=weed (only a few pixels) and transparent=background. Similar visualization for the PhenoBench dataset (c, d).}
\label{fig:figure2}
\end{figure}

\begin{figure}[h]
\centering
    \begin{subfigure}[b]{0.35\linewidth}
        \centering
        \label{fig:sugarbeet_example}
        \includegraphics[width=1\textwidth]{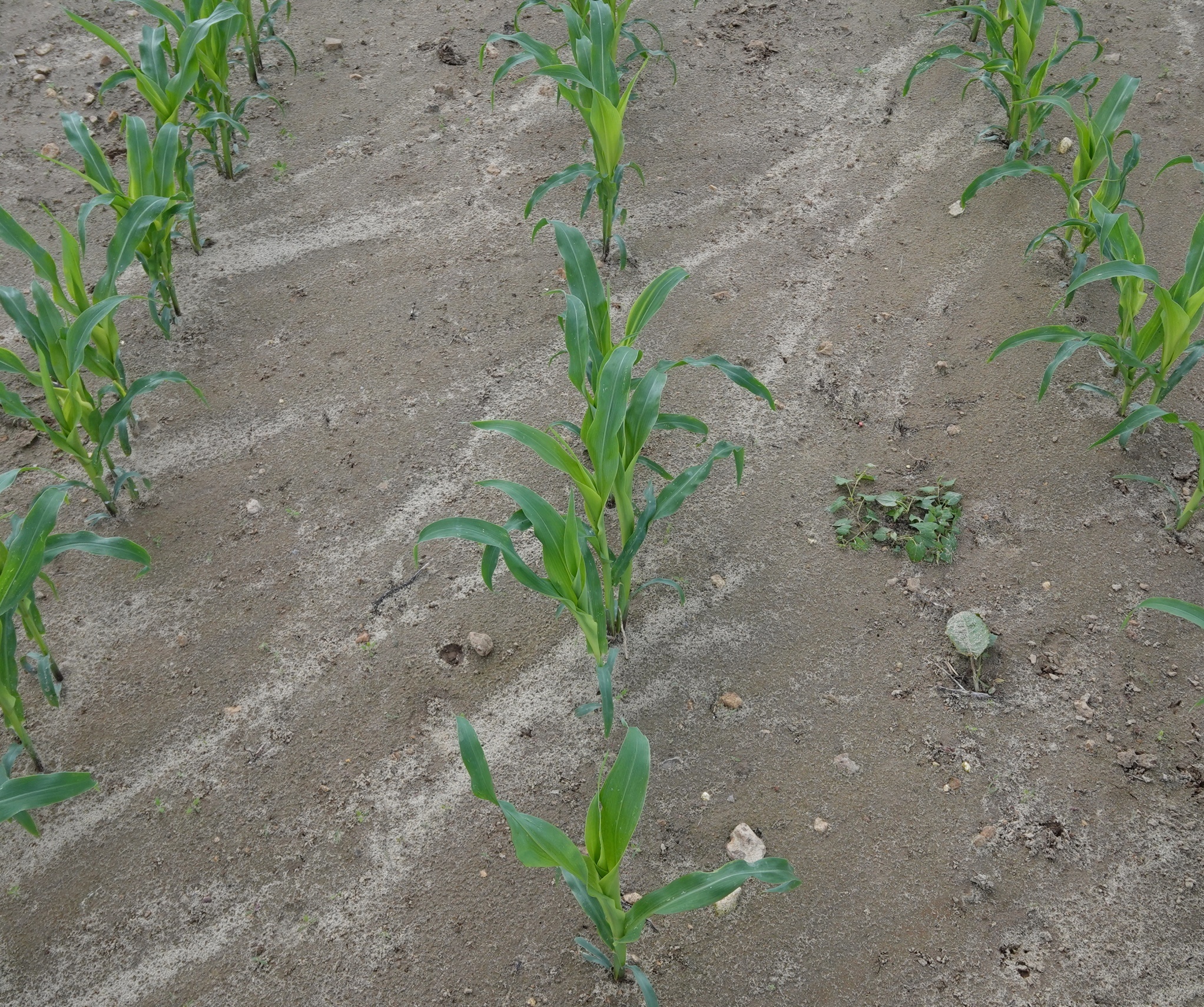}
        \caption[]{}
    \end{subfigure}
    \begin{subfigure}[b]{0.35\linewidth}
        \centering
        \includegraphics[width=1\textwidth]{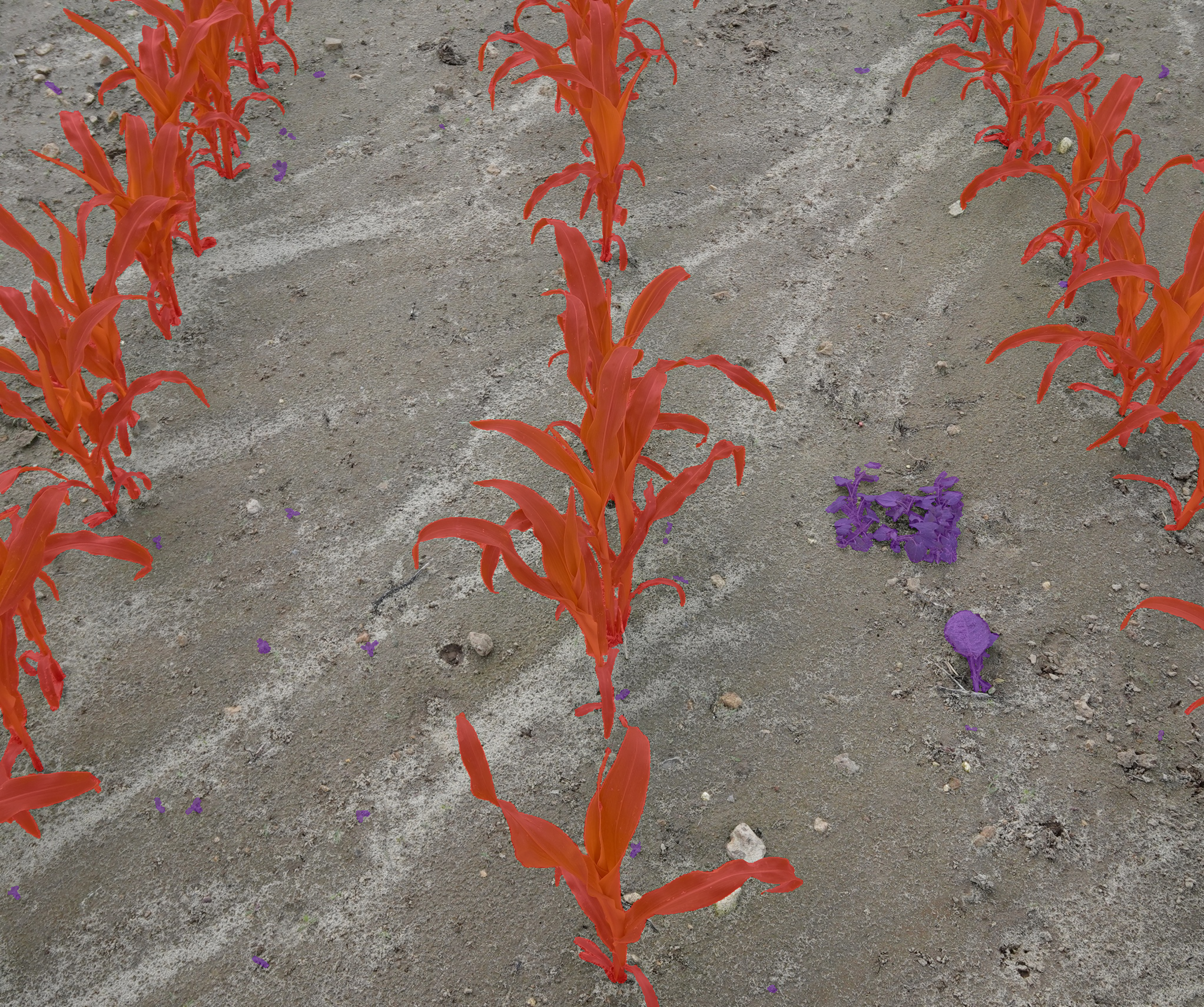}
        \caption[]{}
    \end{subfigure}
    \begin{subfigure}[b]{0.35\linewidth}
        \centering
        \label{fig:sugarbeet_example}
        \includegraphics[width=1\textwidth]{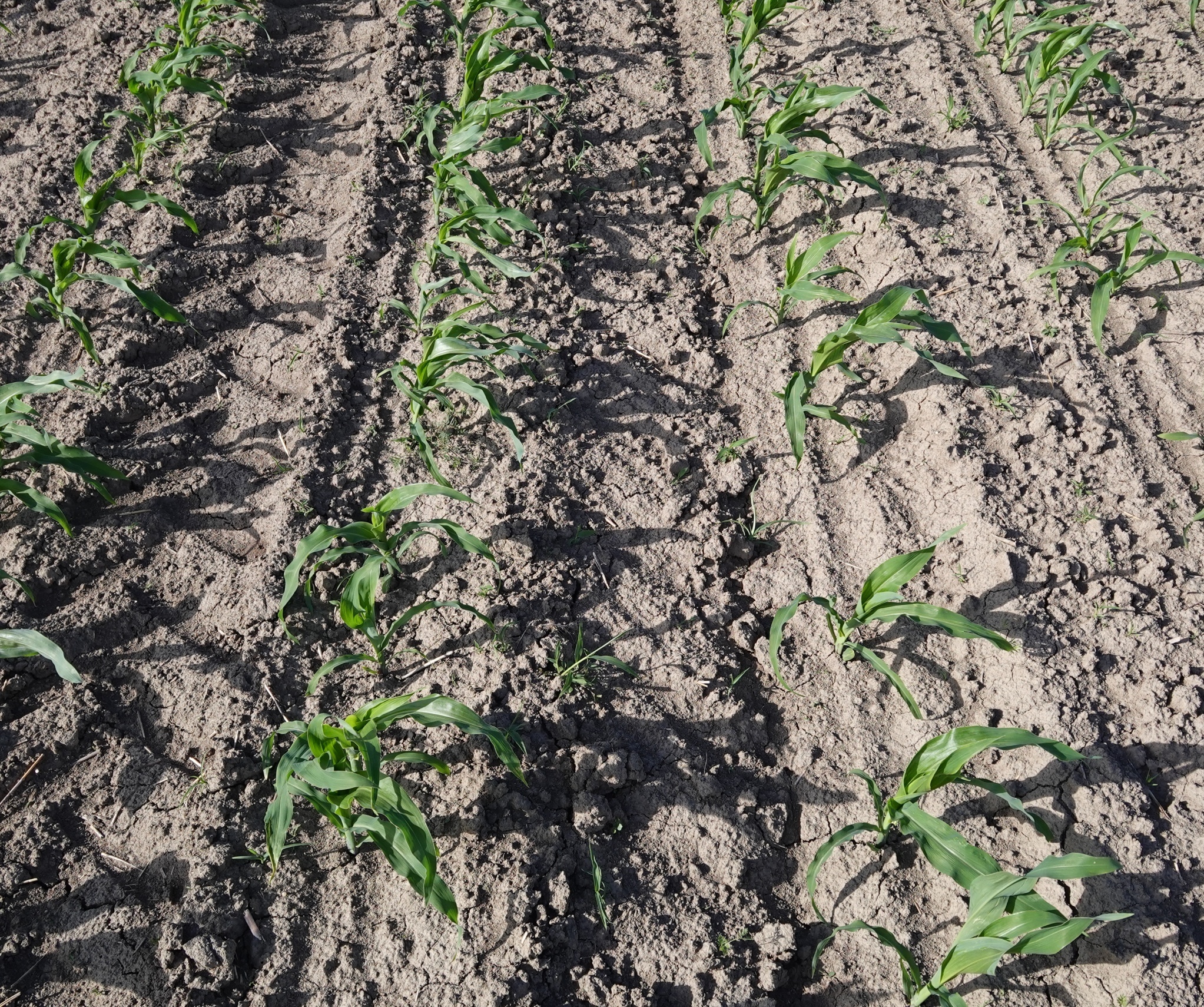}
        \caption[]{}
    \end{subfigure}
    \begin{subfigure}[b]{0.35\linewidth}
        \centering
        \includegraphics[width=1\textwidth]{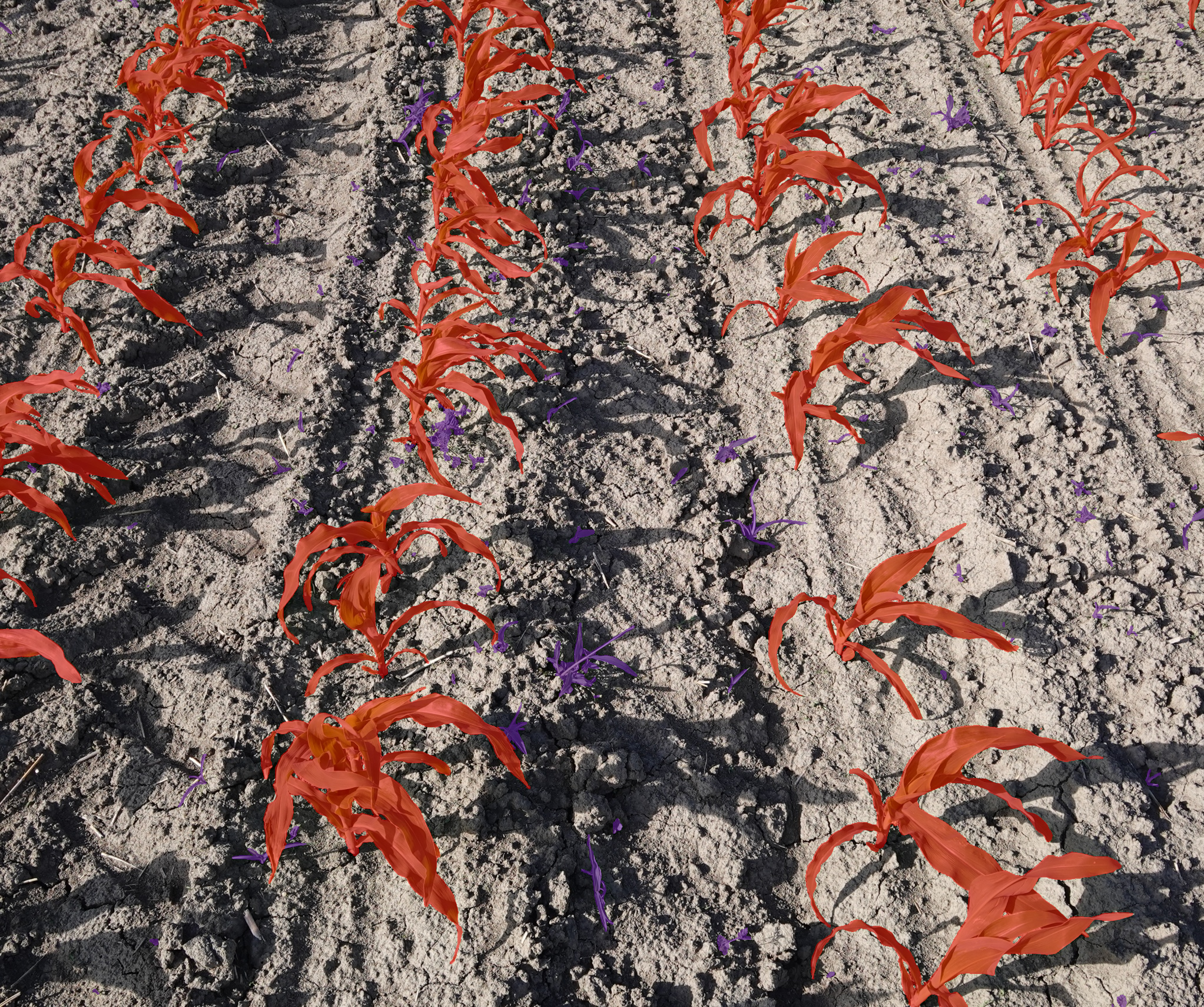}
        \caption[]{}
    \end{subfigure}

\caption{Example image from the Corn-Weed dataset with (a) input image and (b) corresponding annotation, where red=corn, purple=weed (only a few pixels) and transparent=background. Similar visualization for the Corn-Weed pre-trained dataset (c, d).}
\label{fig:corn_dataset}
\end{figure}

In this research the active learning framework was evaluated on two agricultural datasets: "Sugarbeet" and "Corn-Weed". These datasets consisted of images which can be segmented pixel-wise according to three classes: background, weed, and crop (sugarbeet for Sugarbeet, corn for Corn-Weed). The Sugarbeet dataset is a publicly available dataset \cite{chebrolu2017ijrr} that was collected from a single field at multiple acquisition moments during three months. An example image and its ground truth annotation is shown in Figures~\ref{fig:figure2}a and \ref{fig:figure2}b. The Corn-Weed dataset is a private dataset. This dataset is from a production setup representing images from corn fields with weeds. It consisted of images captured at a slight angle as visible in Figures~\ref{fig:corn_dataset}a and \ref{fig:corn_dataset}b. In Table~\ref{tab:dataset_size}, a summary of both datasets is given. Both datasets are heavily imbalanced. Especially the Sugarbeet dataset has a minimal amount of diversity and high image redundancy with 98.5\% of the pixels belonging to the background class. The Corn-Weed is also imbalanced, but it has more pixels related to crop class (6.2\%). Each dataset represents distinctive challenges and conditions corresponding to a variety of agricultural landscapes to which the semantic segmentation model was applied in real-world scenarios. 

%
%

\begin{table}[hbt!]
\centering
\resizebox{\columnwidth}{!}{%
\begin{tabular}{|l|l|l|l|l|l|l|l|}
\hline
Dataset                  & Image resolution     & Train    & Validation & Test   & Background \%  & Crop \%  & Weed \%\\
\hline
Sugarbeet                & 1296 × 966                 & 9287     & 1155       &  1155      & 98.5\%      & 1.3\%      & 0.2\%   \\
Corn-Weed            & 1710 x 2045                 & 331      & 117        &        & 89.8\%      & 6.2\%      & 4.0\%  \\
\hline
\end{tabular}
}
\caption{Active learning dataset statistical summary}
\label{tab:dataset_size}
\end{table}
\vspace{-4mm}

In a production setup, annotated data is often already available. This available data can be used to pre-train a model beforehand to improve uncertainty estimation and active learning sampling (detailed information can be found in Section~\ref{sec:experiments}). Therefore, in this research we have included two additional datasets for pre-training: "PhenoBench" and "Corn-Weed pre-train". The PhenoBench dataset is a publicly available dataset \cite{weyler2023phenobench} with sugarbeets and weeds. Images were acquired on multiple days in two different years, resulting in a dataset with different growth stages and lighting conditions. An example image and its annotation is shown in Figures~\ref{fig:figure2}c and 2d. Corn-Weed pre-train is a diverse dataset consisting of 1190 images from 17 fields spread over 3 countries (Figure~\ref{fig:corn_dataset}c and ~\ref{fig:corn_dataset}d). Compared with the Corn-Weed dataset the acquisition angle is slightly different. In Table~\ref{tab:pretrained_dataset} a summary of both pre-trained datasets are given. In particular, the Phenobench dataset stands out, with 11.8\% of the pixels annotated as crop.

\begin{table}[hbt!]
\centering
\resizebox{\columnwidth}{!}{%
\begin{tabular}{|l|l|l|l|l|l|l|l|}
\hline
Dataset                  & Image resolution   & Train   & Validation & Test   & Background \%  & Crop \%  & Weed \%\\
\hline
PhenoBench               & 1024 x 1024               & 1407     &  772       &        & 87.7\%      & 11.8\%          & 0.5\%   \\
Corn-Weed pre-train       & 1710 x 2045             & 1190     &            &        & 90.9\%      & 5.4 \%      & 3.7 \%  \\
\hline
\end{tabular}
}
\caption{Pre-training dataset statistical summary}
\label{tab:pretrained_dataset}
\end{table}

%
%

\vspace{-8mm}

\section{Experiments}
\label{sec:experiments}
This section describes the experiments conducted on the selected datasets. In all experiments, the active learning process was evaluated based on mean Intersection-over-Union (mIoU). mIoU is a metric for the number of overlapping pixels between the predicted segmentation masks and the ground truth annotations. In this work, the mIoU value represents the average mIoU across all classes (crop, weed, and background).

\subsection{Effect of model pre-training on the active learning performance on SugarBeet}\label{sec:exp_sugarbeet}
Active learning is based on the concept of sampling images about which the model is most uncertain. The concept of uncertainty depends on the knowledge of the model. In the initialization phase of active learning, this knowledge is usually brought in through transfer learning, a process of transferring the weights of the same model trained on a public dataset, such as Cityscapes. These large datasets contain many general classes, such as human and car, but no agricultural classes, such as crop and weed. In this work, a hypothesis is that the uncertainty determination and active learning significantly benefit when transfer learning is done on agricultural images. To test this, the first experiment was tested under two scenarios: 

\begin{enumerate}
    \item Active learning sampling starting with the standard Cityscapes weights. We refer to this in the remainder of this paper as "without agricultural pre-training".
    \item Active learning sampling starting with the weights of a model pre-trained on the PhenoBench (train) dataset \cite{weyler2023phenobench}. We refer to this in the remainder of this paper as "with agricultural pre-training".
\end{enumerate}

For a fair comparison between the two scenarios, the adjustable parameters in active learning were kept the same. A sample size of 10 images was chosen, with number of iterations set to 9; this means that training and sampling were stopped upon reaching 90 images. The dropout probability was set to 0.5. In this experiment, three replicate trials were performed for each acquisition function to account for the randomness in the Monte-Carlo dropout method. A one-way ANOVA with a significance level of 5\% was employed to test whether there were significant performance differences between BALD, PowerBALD and Random. Furthermore, an in-depth analysis was made by making a histogram of the uncertainty values from BALD on the models without and with agricultural pre-training for both the Sugarbeet and Phenobench-validation dataset. If BALD is calculating the uncertainty values accordingly, then images on the Sugarbeet dataset should have a higher uncertainty than the Phenobench-validation images when using agricultural pre-training. The in-depth analysis also contains a visualization of pixel-level uncertainty values for a sample image. Finally, the active learning performance was compared to the benchmark performance by training the segmentation model on the entire training pool (9287 images). The resulting mIoU value was considered as the maximum mIoU that could be achieved on our datasets.

\subsection{Effect of model pre-training on the active learning performance on Corn-Weed}\label{sec:exp_cornweed}



In this experiment, active learning performance was evaluated in an industrial application. Often in agriculture, a model is trained on a set of images obtained from multiple fields. When deploying the model in practice, new fields will be encountered, meaning that the model is exposed to different conditions than the ones learned during model training. This may lead to suboptimal model performance in practice. To overcome this, it is necessary to select and annotate images from new fields, but for humans it is unclear which images should be selected. In this experiment, the goal is to verify the effectiveness of active learning when the segmentation model has already been pre-trained on a dataset with the same camera setup but with different field conditions. To test the added value of the pre-training, the following scenarios were tested: 

\begin{enumerate}
    \item Active learning sampling starting with the standard Cityscapes weights. Again, we refer to this as "without agricultural pre-training".
    \item Active learning sampling starting with the weights of a model pre-trained on the Corn-Weed pre-train dataset of Table~\ref{tab:pretrained_dataset}. Again, we refer to this as "with agricultural pre-training".
\end{enumerate}

In this experiment, the training pool was composed of 331 images from the Corn-Weed dataset (Table~\ref{tab:dataset_size}). The models were tested on the Corn-Weed validation dataset, which consisted of 117 images (Table~\ref{tab:dataset_size}). All active learning settings were similar to Section~\ref{sec:exp_sugarbeet}. The benchmark test was done by training the segmentation model on all 331 images.

\vspace{-5mm}
\section{Results}



\vspace{-2mm}
\subsection{Effect of model pre-training on the active learning performance on SugarBeet}

\begin{figure*}[!h]
        \begin{center}
        \begin{subfigure}[b]{0.49\textwidth}
            \includegraphics[width=\textwidth]{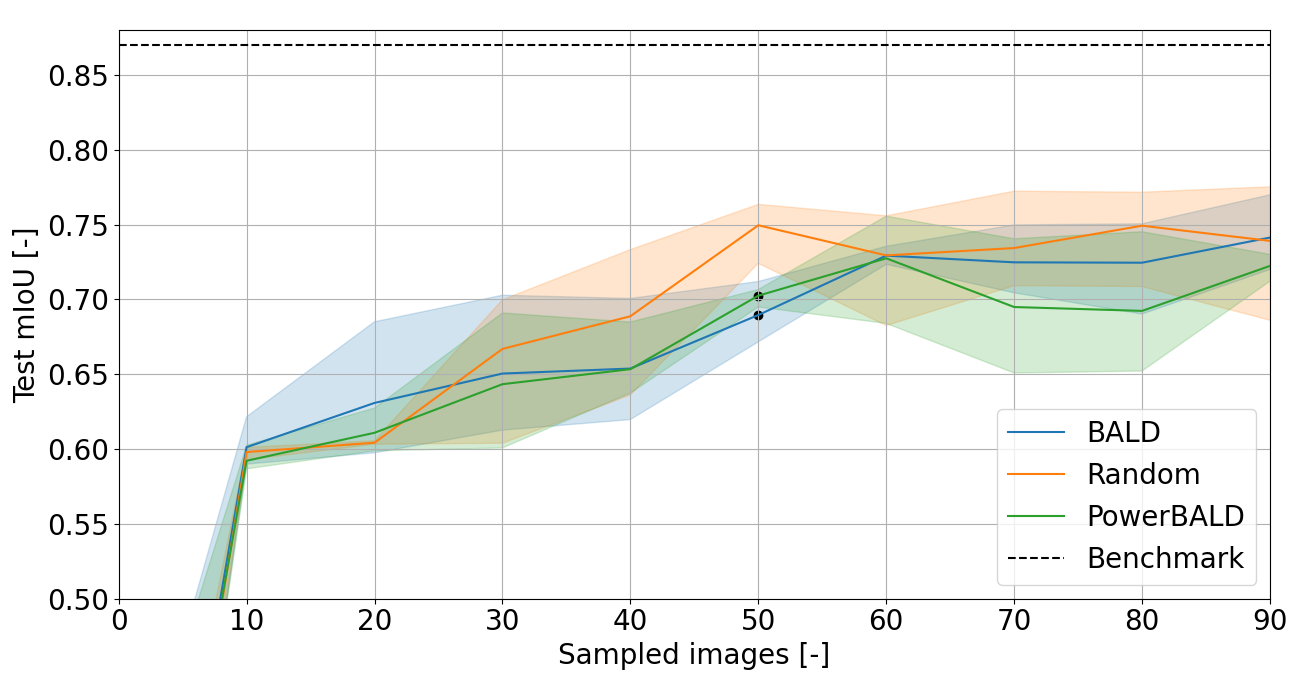}  
            \caption{Sugarbeet without agricultural pre-training}
            \label{fig:sugarbeet_wo_pretraining}
         \end{subfigure}
         \hfill
         \begin{subfigure}[b]{.49\textwidth}
            \includegraphics[width=\textwidth]{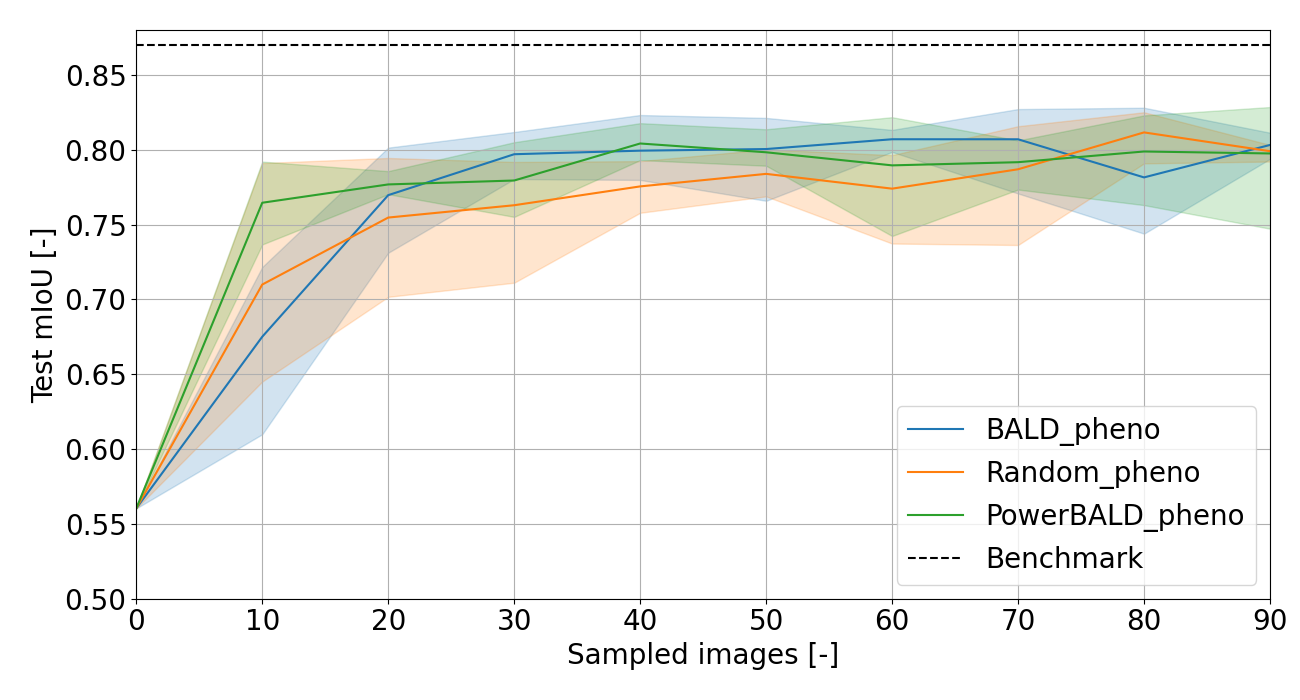}  
            \caption{Sugarbeet pre-trained on PhenoBench}
            \label{fig:sugarbeet_w_pretraining}
         \end{subfigure}
        \end{center}
        \caption{Test performances (mIoU) as a function of the number of training images for BALD, PowerBALD, and Random on the Sugarbeet dataset. Subfigure (a) highlights the results without agricultural pre-training, and subfigure (b) with agricultural pre-training. In both subfigures, the solid colored lines are the mean performance values over the three repetitions. The colored areas around the lines represent the 95\% confidence intervals around the means. The black dashed lines indicate the performance when FCHarDNet was trained on the entire training pool.}
\label{fig:results_sugerbaat}    
\end{figure*}

The experimental results of testing the active learning framework on the Sugarbeet dataset are shown in Figure~\ref{fig:sugarbeet_wo_pretraining} (without agricultural pre-training) and Figure~\ref{fig:sugarbeet_w_pretraining} (with agricultural pre-training). The overall performance of the segmentation model with agricultural pre-training was better compared to the performance without agricultural pre-training. Without agricultural pre-training, the Random acquisition function significantly outperformed (p$\leq$0.05) BALD and PowerBALD at 50 sampled images (see the two black dots in Figure~\ref{fig:sugarbeet_wo_pretraining}). For the other iterations, there were no significant differences between the three acquisition functions. With agricultural pre-training, there were no significant differences observed (Figure~\ref{fig:sugarbeet_w_pretraining}). Despite the lack of significant differences, PowerBALD had a consistently better performance than Random sampling. Furthermore, it should be noted that by only sampling 40 images, the agriculturally pre-trained model had a performance of approximately 0.80 mIoU, which was only 0.07 mIoU lower than the maximum performance of the model trained on all images (black dashed line).


\begin{figure*}[!h]
        \begin{center}
        \begin{subfigure}[b]{0.32\textwidth}
            \includegraphics[width=\textwidth]{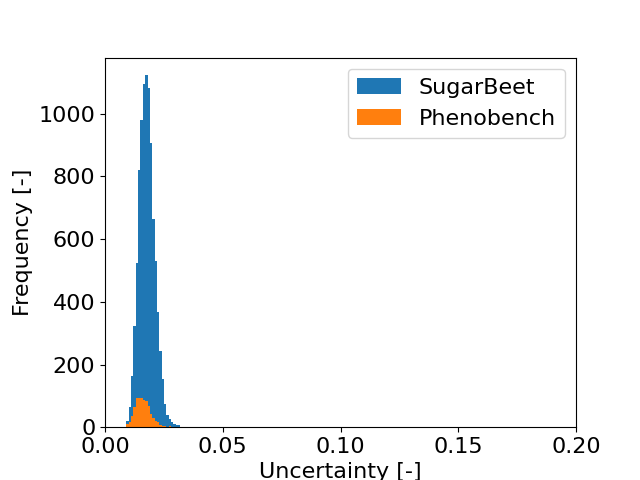}  
            \caption{}
            \label{fig:hist}
         \end{subfigure}
         \hfill
         \begin{subfigure}[b]{0.32\textwidth}
            \includegraphics[width=\textwidth]{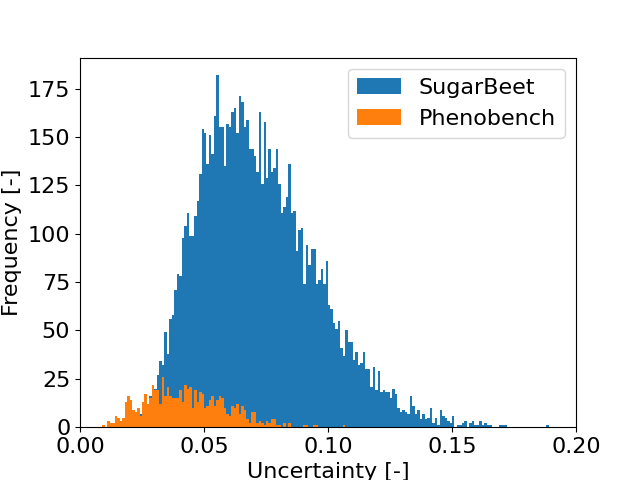}  
            \caption{}
            \label{fig:hist_phenobench}
         \end{subfigure}
         \hfill
         \begin{subfigure}[b]{0.33\textwidth}
            \includegraphics[width=\textwidth]{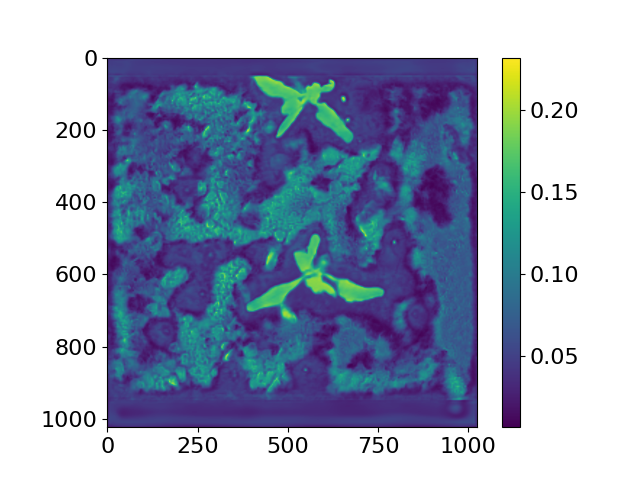}  
            \caption{}
            \label{fig:vis_uncertainty}
         \end{subfigure}
        \end{center}
        \caption{Histogram of the uncertainty values when deploying BALD on a model trained (a) without agricultural pre-training and (b) with agricultural pre-training. In both (a) and (b), the blue histogram depicts the uncertainty values calculated on the 9287 images of the SugarBeet dataset. The orange histogram depicts the uncertainty values calculated on the 772 images of the Phenobench validation dataset. (c) Uncertainty values visualized on a pixel-level in an example image.}
\label{fig:histograms}    
\end{figure*}

The two blue histograms in Figures~\ref{fig:histograms}a and~\ref{fig:histograms}b show the uncertainty values of all images of the training pool when deploying BALD on the segmentation models trained without and with agricultural pre-training. The histogram without pre-training (Figure~\ref{fig:histograms}a) had lower and less scattered uncertainty values than the one with pre-training (Figure~\ref{fig:histograms}b), indicating that BALD was less discriminative in finding uncertain images. This explains the lower performances of both BALD and PowerBALD in the without agricultural pre-training scenario  (Figure~\ref{fig:sugarbeet_wo_pretraining}). The orange histograms in Figures~\ref{fig:histograms}a and~\ref{fig:histograms}b visualize the uncertainty values on the Phenobench-validation dataset. In Figure~\ref{fig:histograms}a, the blue and orange histograms are centered around 0.02, in other words, the images in both validation sets have roughly the same uncertainty. After agricultural pre-training (Figure~\ref{fig:histograms}b), the blue histogram of the SugarBeet dataset has shifted to the right, indicating that images from the SugarBeet dataset got a higher uncertainty value assigned. This is according to our expectation that BALD assigns  higher uncertainty values to images that are relatively unknown to the pre-trained model. Figure~\ref{fig:histograms}c shows the uncertainty values visualized on a pixel-level for an example image of the SugarBeet dataset. The green- and yellow-colored pixels have a higher uncertainty than the blue-colored pixels. Figure~\ref{fig:histograms}c clearly indicates that the model uncertainty was high on the plant pixels.


\subsection{Effect of model pre-training on the active learning performance on Corn-Weed}\label{sec:result_corn}

\begin{figure*}[!h]
        \begin{center}
        \begin{subfigure}[b]{0.49\textwidth}
            \includegraphics[width=\textwidth]{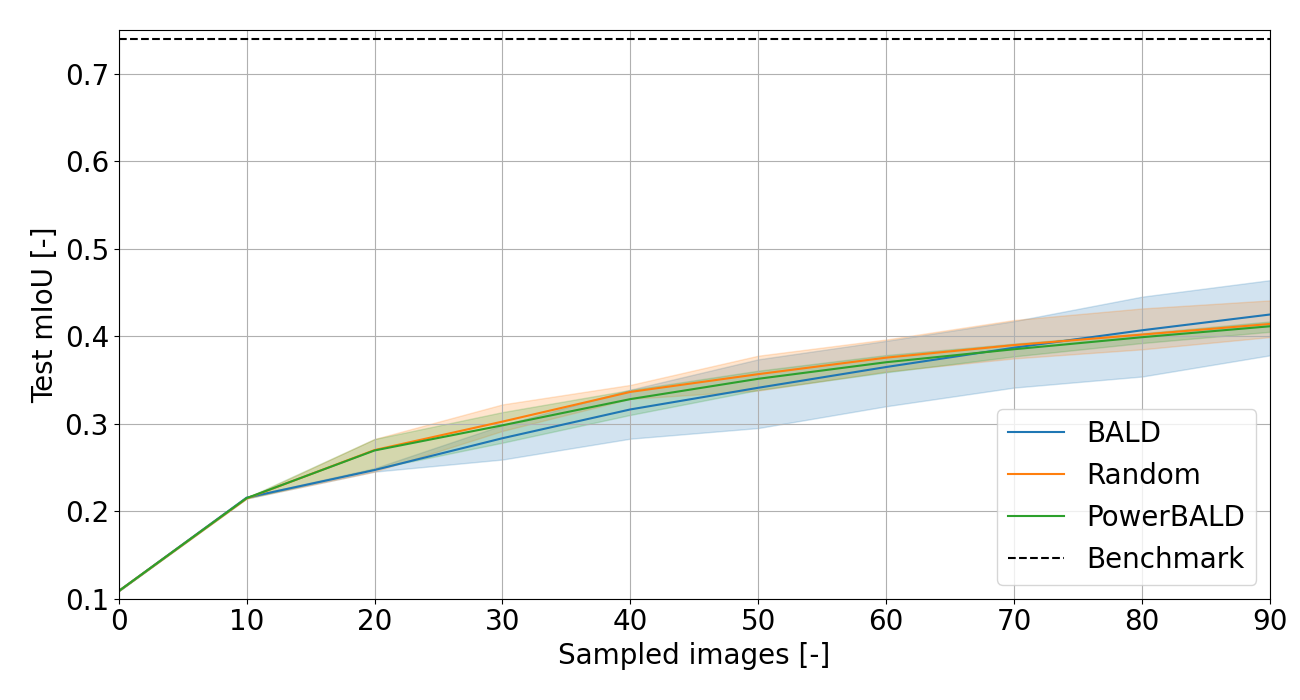}  
            \caption{Corn-Weed without agricultural pre-training}
            \label{fig:low}
         \end{subfigure}
         \hfill
         \begin{subfigure}[b]{0.49\textwidth}
            \includegraphics[width=\textwidth]{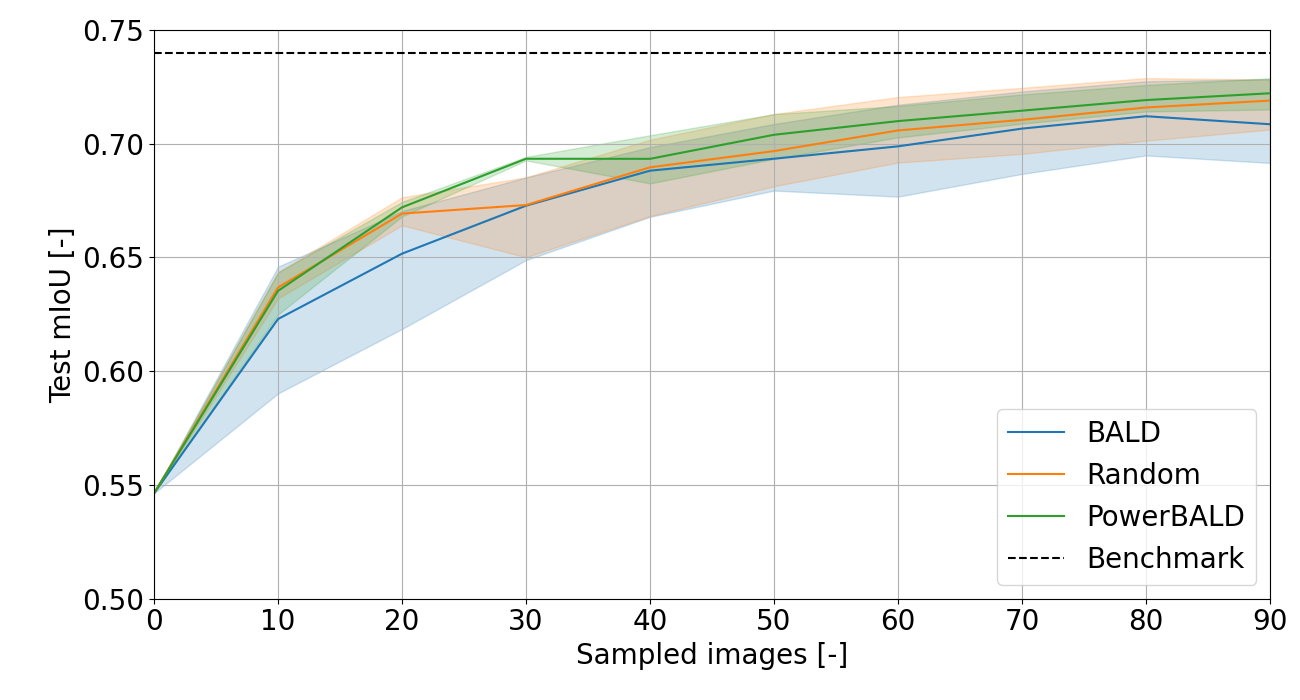}  
            \caption{Corn-Weed with agricultural pre-training}
            \label{fig:medium}
         \end{subfigure}
        \end{center}
        \caption{Test performances (mIoU) as a function of
the number of training images for BALD, PowerBALD, and Random on the Corn-Weed dataset. Subfigure (a) highlights the results without agricultural pre-training, and subfigure (b) with agricultural pre-training. In both subfigures, the solid colored lines are the mean performance values over the three repetitions. The colored areas around the lines represent the 95\% confidence intervals around the means. The black dashed lines indicate the performance when FCHarDNet was trained on the entire training pool.}
\label{fig:corn_result}
\end{figure*}

On the Corn-Weed dataset, we observed marginal differences between the three acquisition functions. Without agricultural pre-training (Figure~\ref{fig:low}), the performances were very similar and not significantly different. With agricultural pre-training (Figure~\ref{fig:medium}), there were also no significant differences between PowerBALD, BALD, and Random. This was mainly due to the high standard deviations between the repetitions of the Random acquisition (see the blue areas in Figure~\ref{fig:medium}). Interestingly, PowerBALD had the lowest variance between the three repetitions (Figure~\ref{fig:medium}). This indicates that PowerBALD had a more consistent performance than BALD and Random on the Corn-Weed dataset. This result is promising when it comes to applying active learning in a production setup. BALD was on average underperforming compared to Random and PowerBALD, meaning that selecting images with only the highest uncertainty did not lead to better model performance on the unseen corn field.

\vspace{-8mm}
\section{Discussion}

In this research, the active learning framework was tested on two agricultural datasets with three acquisition functions: BALD, PowerBALD and Random. On the Sugarbeet dataset, both BALD and PowerBALD did not show a significantly better performance than Random acquisition. There are two possible explanations for the minimal difference between the acquisition functions. 
First, there was a high image redundancy in the Sugarbeet dataset. Training the model on the complete training pool (9287 images) resulted in a mIoU of 0.87. In Figure~\ref{fig:sugarbeet_w_pretraining}, it was shown that after 4 iterations (40 sampled images), the model had a mIoU of approximately 0.80. This indicates that by training on only 40 images, a performance can be obtained that is close to the performance of training on 9287 images. This implies that there was not much image diversity in the dataset. 

The second reason for the minimal difference is that BALD's epistemic uncertainty was calculated across all semantic classes. The fact that the background class represented more than 98\% of the pixels means that BALD probably placed more emphasis on optimizing the background class than on the crop and weed classes. This is also why active learning has been successfully applied to instance segmentation in agriculture \cite{blok2022active}, because there the uncertainty calculation was conducted on the instance level, ignoring the background pixels. To solve the problem of over-present but less relevant classes in semantic segmentation, it may be worthwhile to calculate the sum of uncertainties for the top-K most uncertain pixels. By focusing on the uncertainty sum of a small part of the image, the influence of the background class can be reduced. This would work for the image shown in Figure~\ref{fig:vis_uncertainty}, because all pixels related to the plant had a high uncertainty. Another possibility is to test other acquisition functions, such as Entropy. In a recently published article, it was shown that Entropy was effective for quantifying uncertainty on the Sugarbeet dataset \cite{celikkan2023semantic}. In their work, high pixel uncertainties were observed around the edges of the plants. However, it should be noted that in their research the algorithm was pre-trained on 75\% of the dataset. It would be interesting to see if uncertain pixels are observed when testing on a larger set of unseen images. 


%

The results on the Corn-Weed dataset showed that PowerBALD was better than Random sampling  (Section~\ref{sec:result_corn}). The performance of BALD was worse than Random acquisition. To verify this outcome, additional experiments were done on Cityscapes. In these experiments both PowerBALD and BALD were outperforming Random, indicating that these acquisition functions can be used for active learning. The high redundancy in our agricultural datasets probably made it more difficult for the active learning to outperform the Random acquisition. In our experiments, selection was performed on images that differed from those used in pre-training the model. For example, images in the Corn-Weed dataset had fewer shadows and a more uniform sandy soil than the images in Corn-Weed pre-trained dataset (Figure \ref{fig:corn_dataset}a versus \ref{fig:corn_dataset}c). Due to the difference in images between the two datasets, each randomly sampled image is still relatively high-informative for the segmentation network. In the situation that the training pool also contained images similar to the pre-trained dataset, then there would be a higher probability of selecting non-informative images when using Random acquisition. In that scenario, we would expect larger performance differences between the Random acquisition and the uncertainty-based acquisition functions. Figure~\ref{fig:hist_phenobench} highlights this scenario because the blue histogram had higher uncertainties than the orange histogram. Unfortunately, in our experiments with the relatively small sample sizes and iterations, there was no clear performance difference between the active learning and the Random acquisition. To investigate the added value of active learning in a future research, it is recommended to do additional testing with different numbers of sample sizes, iterations and by including more similar or redundant images in the training pool. The latter would be a better representation of real agricultural datasets, especially when those are obtained with slow-moving platforms.

In future work, our active learning framework could also be improved. Our proposed framework now uses Monte-Carlo dropout, which requires the same image to be analyzed multiple times to be able to calculate the model uncertainty. The disadvantage of Monte-Carlo dropout is that it takes significantly more time to conduct the image analysis compared to a standard image analysis. Therefore, Monte-Carlo dropout is not preferred for larger datasets, because of its limited scalability \cite{sener2017active}. In future work, other potentially more time-efficient sampling methods can be tested, including the learning loss method \cite{yoo2019learning}. With learning loss, an additional regression module is added to the neural network architecture to predict the loss of unlabeled images. The advantage of this method is that it can predict the image uncertainty in one forward pass, thus drastically reducing the sampling time compared to Monte-Carlo dropout sampling. Next to that, active learning was evaluated by comparing it to Random sampling. In industrial applications, pre-selection of fields or locations is often done by humans. In the research of \cite{Chandra2020}, human annotators decided which images should be annotated. Since humans are often involved in selection tasks, it may be interesting to compare the active learning with human sampling as well.  



\section{Conclusions}
Active learning focuses on optimizing model performance with fewer image annotations. In this work, PowerBALD was shown to outperform Random acquisition when the segmentation model was pre-trained on an agricultural dataset. PowerBALD also showed less variation in performance across the three repetitions than Random acquisition, indicating better sampling stability. Although the better trend, there were no significant differences observed (p$\leq$0.05). The lack of other significant differences can be explained by the fact that the datasets were highly unbalanced with more than 89\% of the pixels belonging to the less relevant background class. Future research should focus on sampling the top-K most uncertain pixels to address the class imbalance at image level.

\section*{Acknowledgments}
We would like to thank the following people from Wageningen University and Research, who helped us in our research: Ard Nieuwenhuizen and Trim Bresilla.


\printbibliography

\end{document}